\documentclass{article}

\usepackage[preprint]{neurips_2026}
 \usepackage{amsmath}
 \usepackage{subcaption}
 \usepackage{comment}
\usepackage{float}
\usepackage[utf8]{inputenc} 
\usepackage[T1]{fontenc}    
\usepackage{hyperref}       
\usepackage{url}            
\usepackage{booktabs}       
\usepackage{amsfonts}       
\usepackage{nicefrac}       
\usepackage{microtype}      
\usepackage{xcolor}         
\usepackage{natbib}
\usepackage{graphicx}
 \usepackage{booktabs} 
 \usepackage{multirow}
  \usepackage{amssymb}

\title{Sanity Checks for Long-Form Hallucination Detection}

\author{%
  Geigh Zollicoffer \\
  Los Alamos National Laboratory \\
  \And
  Minh Vu \\
  Los Alamos National Laboratory \\
  \And
  Hongli Zhan \\
  The University of Texas at Austin \\
  \And
  Raymond Li \\
  University of British Columbia \\
  \And
  Manish Bhattarai \\
  Los Alamos National Laboratory \\
}

\begin{document}

\maketitle

\begin{abstract}
Hallucination detection methods for large language models increasingly operate on chain-of-thought reasoning traces, yet it remains unclear whether they evaluate the reasoning itself or merely exploit surface correlates of the final answer.
We introduce a controlled-invariance methodology that exposes this distinction through two oracle tests: \textsc{Force}, which replaces each response's final answer with the ground truth while preserving the reasoning trace, and \textsc{Remove}, which strips answer-announcement steps while leaving the trajectory intact.
This reveals if their predictive power derives from answer-level artifacts rather than from the structure or validity of intermediate reasoning.
We further show that once these artifacts are controlled for, effective detection does not necessarily require complex learned representations: TRACT, a lightweight scorer built on lexical trajectory features (hedging trends, step-length dynamics, and cross-response vocabulary convergence), achieves strong robustness while remaining competitive with or outperforming existing baselines on unperturbed traces.
These findings suggest that the current central challenge in reasoning-aware hallucination detection is not the absence of signal in the trace, but the failure to isolate it from endpoint cues.
\end{abstract}

\section{Introduction}

As large language models (LLMs) are increasingly used for reasoning and decision support, their reliability depends on detecting when generated outputs are unsupported, inconsistent, or false.
Hallucination detection addresses this problem, but a central question remains unresolved: do current detectors assess the reasoning process itself, or do they mainly exploit surface correlates of the final answer?

This question is especially important for long-form reasoning.
Recent hallucination and uncertainty detectors increasingly operate on chain-of-thought traces, comparing sampled responses, measuring semantic agreement, or scoring reasoning-path consistency~\citep{Wang_Su_Ai_Liu_2026}.
Yet strong performance on unmodified traces does not by itself show that a detector is reasoning-faithful.
A method may appear to evaluate intermediate reasoning while actually relying on endpoint cues, answer formatting, response length, or coarse agreement among final answers.
In that case, reported gains may overstate progress toward genuine reasoning assessment.

\begin{figure}[!ht]
  \centering
  \includegraphics[width=.85\linewidth]{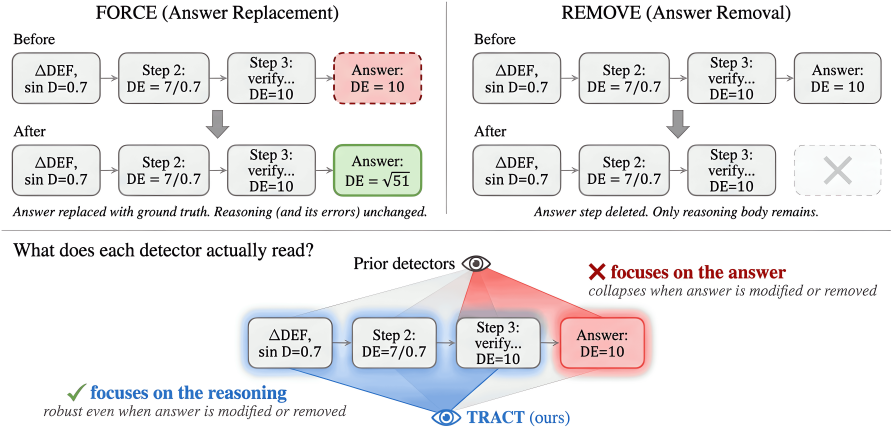}
  \caption{\textbf{Two sanity-check operations.}
\textsc{Force} replaces only the final answer with the ground truth; \textsc{Remove} deletes explicit answer-announcement steps.
Both preserve the reasoning body, so a trace-faithful detector should remain informative.}
  \label{fig:teaser}
\end{figure}

We introduce a controlled-invariance framework for exposing this failure mode.
The key idea is simple: if a detector claims to evaluate the reasoning trajectory, then transformations that preserve the reasoning body should not destroy its ability to distinguish correct from incorrect reasoning.
We instantiate this idea with two oracle sanity checks, illustrated in Figure~\ref{fig:teaser}.
In \textsc{Force}, we replace the final answer with the ground truth and canonicalize its presentation, while leaving the intermediate reasoning unchanged.
In \textsc{Remove}, we delete explicit answer-announcement steps, again preserving the reasoning body.
Neither intervention repairs a flawed derivation or corrupts a valid one.
Thus, a trace-faithful detector should remain informative under both conditions; large shifts indicate dependence on answer-level artifacts rather than reasoning evidence.

\begin{figure}[!ht]
    \centering
    \includegraphics[width=.85\linewidth]{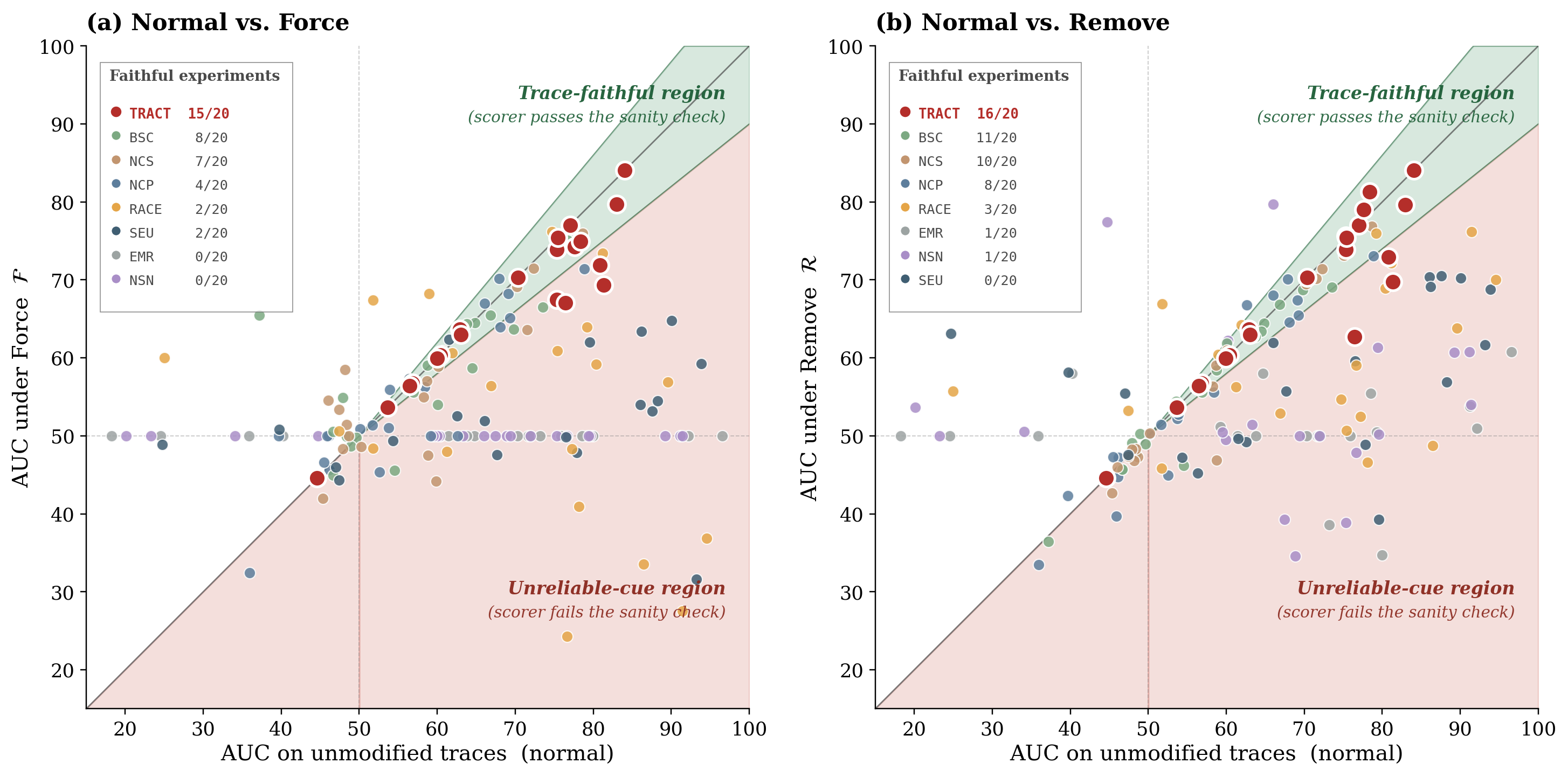}
\caption{\textbf{Sanity-check results across four benchmarks and five models.}
Each point is one scorer--model--benchmark experiment; $x$ is AUC on original traces and $y$ is AUC after \textsc{Force} or \textsc{Remove}.
Trace-faithful scorers should remain near the diagonal because the reasoning body is preserved.
TRACT has the highest number of faithful settings under both interventions.}
    \label{fig:stability_plot}
\end{figure}

Applying these tests across four benchmarks and five models reveals that many existing detectors are less trace-faithful than standard evaluations suggest.
As shown in Figure~\ref{fig:stability_plot}, several methods move far from the diagonal under \textsc{Force} or \textsc{Remove}, meaning their discriminative behavior changes substantially even though the intermediate reasoning trajectory is preserved.
This is not merely a calibration issue: it indicates that some detectors obtain much of their signal from endpoint availability, answer standardization, or other artifacts that are orthogonal to reasoning quality.

We then ask whether robust trace-level detection requires complex learned representations.
Surprisingly, it does not.
We propose TRACT, a lightweight black-box scorer built from lexical trajectory features: local coherence cues, structural dynamics such as hedge and step-length trends, and cross-sample content convergence.
TRACT is not a proof checker and does not verify each intermediate step.
Instead, it targets observable symptoms of unsettled reasoning: traces that wander, hedge increasingly, become structurally irregular, or fail to converge across independent samples.
Because these features are computed from the reasoning body rather than the endpoint string, TRACT is naturally suited to the \textsc{Force}/\textsc{Remove} setting.

Our results support two conclusions.
First, oracle robustness testing is a necessary sanity check for reasoning-aware hallucination detection: without it, a detector may appear successful while relying on answer-level artifacts.
Second, useful trace-level signal exists in simple, interpretable trajectory statistics.
The challenge is therefore not only to design stronger detectors, but to evaluate whether their strength comes from the reasoning process they are meant to assess.

\section{Background and Related Work}
\label{sec:background_related_work}

\paragraph{Sampling-based uncertainty.}
A common black-box approach to hallucination detection is to sample multiple responses and measure their disagreement: models that know the answer should produce mutually consistent outputs, while hallucinated or uncertain generations tend to diverge.
Semantic entropy~\citep{farquhar2024semantic,kuhn2023semantic} formalizes this idea by clustering meaning-equivalent responses and computing entropy over the resulting semantic classes.
Subsequent work refines either the representation or the disagreement measure: Kernel Language Entropy replaces hard clustering with a continuous similarity kernel~\citep{nikitin2024kernel}, while embedding-based methods such as SINdex~\citep{sindex} and Semantic Embedding Uncertainty (SEU)~\citep{grewal2024improvinguncertaintyquantificationlarge} use dense sentence representations to estimate inconsistency more efficiently.
Perturbation-based variants further sample over input transformations rather than only model randomness~\citep{gao2024spuq}.
These methods provide strong black-box uncertainty signals, but their scores are usually computed over complete responses or final-answer semantics, making it difficult to tell whether they measure reasoning quality or endpoint agreement.

\paragraph{Reasoning-trace-aware detection.}
Chain-of-thought prompting~\citep{wei2022chainofthought} and long-form reasoning models make the intermediate trajectory observable, motivating detectors that evaluate not only what answer is produced but how the answer is reached.
RACE~\citep{Wang_Su_Ai_Liu_2026} is representative of this direction, combining inter-sample reasoning-path consistency, answer uncertainty, reasoning--answer alignment, and intra-trace coherence.
Such methods are closer to reasoning-aware hallucination detection than answer-only uncertainty estimators, but they also introduce a new evaluation problem: high performance may still come from endpoint cues, answer alignment, or coarse consistency rather than from trace-faithful assessment of the reasoning body.
Our \textsc{Force} and \textsc{Remove} tests are designed to expose this distinction.

\paragraph{Unified UQ frameworks and baselines.}
Recent unified frameworks collect many of these signals into calibrated uncertainty pipelines.
For example, \textsc{uqlm}~\citep{JMLR:v27:25-1557} includes exact-match repetition and diversity scores~\citep{cole2023selectively}, $n$-gram and BERTScore self-consistency~\citep{manakul2023selfcheckgpt,zhang2020bertscore}, NLI-based non-contradiction probability~\citep{chen2023quantifying}, sentence-embedding similarity~\citep{reimers2019sentencebert}, and semantic entropy variants~\citep{farquhar2024semantic,kuhn2023semantic}.
Surveys similarly emphasize uncertainty quantification as a central route to improving LLM reliability, while highlighting trade-offs among accuracy, cost, access requirements, and interpretability~\citep{shorinwa2025survey,kang2025uqsurvey}.
We use these families as black-box baselines because they cover the main operational signals in current hallucination detection: answer repetition, lexical overlap, embedding similarity, NLI agreement, semantic entropy, and reasoning-path consistency.

\paragraph{White-box versus black-box access.}
White-box detectors use token probabilities or hidden-state activations to estimate reliability in a single pass~\citep{duan2024shifting,zollicoffer2025mtremultitokenreliabilityestimation,phukan-etal-2025-beyond,binkowski-etal-2025-hallucination,fadeeva2024factchecking}.
These methods can be effective, but require model internals that are unavailable for many closed-source systems.
We therefore focus on the black-box setting, where the detector observes only sampled text traces.
Within this setting, our goal is not merely to improve AUC, but to test whether a detector's signal remains valid when answer-level artifacts are controlled.

\section{TRACT: Trace Rhetorical and Coherence Trajectory}
\label{sec:tract}

A correct reasoning trace tends to \emph{settle}.
As the model approaches a solution, its steps usually become more directed: the vocabulary stabilises, intermediate claims become more consistent, and independently sampled traces begin to resemble one another in how they progress and where they end.
By contrast, an incorrect trace often \emph{wanders}: it asks itself unnecessary questions, restarts or hedges, expands when it should compress, and disagrees with parallel samples about how many steps the problem requires.

TRACT operationalises this observation in a fully black-box sampling setting.
Given a prompt $x$, we sample $K$ independent reasoning traces from the model,
$\{\mathbf{r}^{(k)}\}_{k=1}^{K}$.
Each trace is written as a sequence of textual steps,
$\mathbf{r}^{(k)} = (s^{(k)}_1,\ldots,s^{(k)}_{T_k})$.
A textual step is the smallest reasoning unit exposed in the response, such as a numbered line, bullet point, sentence-level inference, or explicitly separated intermediate statement.
For example, in a chain-of-thought response, ``First, compute the total cost'' and ``Therefore, the remaining amount is 12'' would be treated as two separate steps.
Answer-announcement steps such as ``Final Answer:'' are excluded before TRACT features are computed, so the scorer operates on the reasoning body rather than the endpoint string.

TRACT does not require access to logits, hidden states, answer labels, embedding models, or auxiliary entailment models.
Instead, it reads the sampled traces themselves and extracts lightweight trajectory-level features that describe how the reasoning behaves.
The features are grouped into three blocks, each corresponding to a different diagnostic question:
i) \textbf{Coherence:} what does each trace look like on average?
ii) \textbf{Structure:} how does the trace evolve as reasoning progresses?
iii) \textbf{Content:} do independent traces converge on the same intermediate and final vocabulary?

Table~\ref{tab:tract_features} summarises the full TRACT feature set.
The table is organised around the observable signature of each feature: what pattern it detects, whether that pattern increases the hallucination score, and how the statistic is computed.
The prose below explains why these signatures are useful.

\begin{table*}[t]
\centering
\small
\setlength{\tabcolsep}{5pt}
\renewcommand{\arraystretch}{1.18}
\resizebox{\textwidth}{!}{%
\begin{tabular}{@{}p{0.13\linewidth} l p{0.34\linewidth} c p{0.36\linewidth}@{}}
\toprule
\textbf{Block} & \textbf{Feature} & \textbf{Intuition} & \textbf{Sign} & \textbf{Definition} \\
\midrule

\multirow{3}{*}{\parbox{0.13\linewidth}{\textbf{Coherence}\\$\phi_{\text{coh}}$}}
  & \textsc{QuestionRate}
  & Frequent self-questioning indicates unresolved uncertainty.
  & $+$
  & $\frac{1}{K}\sum_k \frac{1}{T_k}\sum_i q^{(k)}_i$ \\

  & \textsc{WordsPerStep}
  & Verbose steps often reflect hedging, qualification, or failure to compress.
  & $+$
  & $\frac{1}{K}\sum_k \frac{1}{T_k}\sum_i w^{(k)}_i$ \\

  & \textsc{PlateauFrac}
  & Non-growing steps suggest the trace is stalling rather than developing.
  & $+$
  & $\frac{1}{K}\sum_k \frac{1}{T_k-1}\sum_{i=2}^{T_k}\mathbb{I}\!\left[w^{(k)}_i \le w^{(k)}_{i-1}\right]$ \\

\midrule

\multirow{5}{*}{\parbox{0.13\linewidth}{\textbf{Structure}\\$\phi_{\text{struct}}$}}
  & \textsc{HedgeSlope}
  & Increasing hedge density suggests confidence is draining over time.
  & $+$
  & $\frac{1}{K}\sum_k \mathrm{slope}_{i/T_k}\!\left(h^{(k)}_i\right)_{i=1}^{T_k}$\\

  & \textsc{ColonFrac}
  & Explicit delimiters mark organised decomposition, cases, or subclaims.
  & $-$
  & $\frac{1}{K}\sum_k \frac{1}{T_k}\sum_i \mathbb{I}\!\left[\texttt{:} \in s^{(k)}_i\right]$ \\

  & \textsc{MaxStepWc}
  & A focused long step can indicate concentrated reasoning depth.
  & $-$
  & $\frac{1}{K}\sum_k \max_i w^{(k)}_i$ \\

  & \textsc{ScMax}
  & An outlier-length trace signals disagreement about the required reasoning path.
  & $+$
  & $\max_k T_k$ \\

  & \textsc{WcVarSlope}
  & Growing local length variance suggests the trace is losing its structural thread.
  & $+$
  & $\frac{1}{K}\sum_k \mathrm{slope}_{i/T_k}\!\left(\mathrm{Var}(w^{(k)}_{i-2:i})\right)_{i=3}^{T_k}$ \\

\midrule

\multirow{3}{*}{\parbox{0.13\linewidth}{\textbf{Content}\\$\phi_{\text{cont}}$}}
  & \textsc{MidUnigramDiv}
  & Mid-trace lexical disagreement indicates early divergence across samples.
  & $+$
  & $\binom{K}{2}^{-1}\sum_{j<k}\left(1-J(\mathcal{U}^{(j)}_{m_j},\mathcal{U}^{(k)}_{m_k})\right)$ \\

  & \textsc{FinalUnigramDiv}
  & Terminal lexical disagreement indicates that independent traces end in different places.
  & $+$
  & $\binom{K}{2}^{-1}\sum_{j<k}\left(1-J(\mathcal{U}^{(j)}_{T_j},\mathcal{U}^{(k)}_{T_k})\right)$ \\

  & \textsc{EntityRepeat}
  & Repeatedly revisiting the same entity can indicate circular reasoning.
  & $+$
  & $\frac{1}{K}\sum_k \frac{1}{T_k}\sum_{i=2}^{T_k}\mathbb{I}\!\left[\mathcal{E}^{(k)}_i \cap \mathcal{E}^{(k)}_{i-1} \neq \varnothing\right]$ \\

\bottomrule
\end{tabular}%
}
\caption{\textbf{TRACT feature inventory.}
TRACT maps sampled reasoning traces into three interpretable feature blocks.
The intuition column gives the diagnostic signature; the definition column gives the implementation-level statistic.
Here $T_k$ is the number of steps in trace $k$, $s^{(k)}_i$ is step $i$, $w^{(k)}_i=|s^{(k)}_i|$ is its word count, $q^{(k)}_i$ is its question-mark count, $h^{(k)}_i$ is the number of hedge words from lexicon $\mathcal{H}$ in the step, $\mathcal{U}^{(k)}_i$ is the lowercased unigram set, $\mathcal{E}^{(k)}_i$ is the set of capitalised tokens, $m_k=\lfloor T_k/2\rfloor$, and $J(A,B)=|A\cap B|/|A\cup B|$ is Jaccard similarity. Sign indicates whether larger feature values increase $(+)$ or decrease $(-)$ the TRACT incorrectness score.}
\label{tab:tract_features}
\end{table*}

\paragraph{Coherence features: does each trace sound locally resolved?}

The coherence block captures whether a trace has the local rhetorical profile of a settled solution.
When a model knows how to proceed, its steps tend to state intermediate claims directly and move on.
When it is uncertain, the trace often becomes interrogative, over-qualified, or repetitive.
\textsc{QuestionRate} captures explicit self-questioning, \textsc{WordsPerStep} captures verbosity and over-explanation, and \textsc{PlateauFrac} captures steps that fail to develop.

This block helps because many reasoning failures are visible before the final answer: the model circles around the problem, asks itself what to do, or spends words compensating for a missing solution path.
Coherence features are therefore weak but useful symptoms of local unresolvedness.

\paragraph{Structure features: does the trace keep its trajectory?}

The structure block captures whether reasoning becomes more organised or more unstable over time.
A correct trace can be long or short, but it usually maintains a coherent trajectory: decomposition, intermediate work, and convergence.
An incorrect trace more often loses this trajectory.
Hedging may increase, step lengths may become irregular, or one sampled trace may become much longer than the others because the model cannot settle on a path.

TRACT measures this with five structural signatures.
\textsc{HedgeSlope} tracks whether uncertainty language grows over ~\citep{lakoff1973hedges,katerenchuk2024youprobablyreadthis}.
\textsc{ColonFrac} captures explicit organisation through cases, lists, or subclaims.
\textsc{MaxStepWc} captures whether the trace contains a concentrated reasoning step.
\textsc{ScMax} flags an outlier-length trace among the $K$ samples.
\textsc{WcVarSlope} tracks whether the rhythm of step lengths becomes increasingly unstable~\citep{jin-etal-2024-impact}.

This block helps because reasoning quality is often dynamic: the important signal is not only what the trace contains, but how it changes~\citep{vanhoyweghen2025lexicalhintsaccuracyllm}.

\paragraph{Content features: do independent traces converge?}

The content block captures whether independent samples appear to reason toward the same state.
Correct traces need not be identical, but they often share key concepts at the midpoint and converge to similar terminal vocabulary.
Incorrect traces are more likely to scatter: one sample pursues one interpretation, another pursues a different one, and their intermediate or final steps share little lexical overlap.

TRACT measures this with lightweight lexical agreement.
\textsc{MidUnigramDiv} asks whether traces agree around the middle of the reasoning process.
\textsc{FinalUnigramDiv} asks whether they agree near the end.
\textsc{EntityRepeat} captures a different content failure: a single trace repeatedly revisiting the same named entity instead of advancing the reasoning state~\cite{yao-etal-2025-understanding,duan2026circular}.
This block helps because uncertainty is often expressed across samples.
Even when each individual trace looks fluent, disagreement among independently sampled trajectories can reveal that the model has not identified a stable solution.

\paragraph{Scoring}

TRACT maps the three feature blocks into a single incorrectness score.
Each feature is robust-scaled using median centring and IQR normalisation, and then clipped to $[-3,3]$ to limit the effect of extreme traces.
Within each block, features use fixed equal-magnitude weights, with signs given in Table~\ref{tab:tract_features}. 
The structure block always contributes to the score.
This block captures trajectory dynamics---hedge trends, step-count outliers, and step-length irregularity---that remain meaningful across both terse and verbose reasoning styles.
The coherence and content blocks are more style-dependent.
They are most reliable when responses are explicitly segmented into step-wise reasoning; in prose-heavy traces, the same surface cue can change meaning.
For example, self-questioning may signal confusion in terse chain-of-thought, but careful exposition in a long explanatory paragraph.
TRACT therefore gates the coherence and content blocks using the raw \textsc{WordsPerStep} value.

Formally, TRACT computes
\[
    \mathrm{TRACT}(x)
    =
    \mathbf{w}_{\mathrm{struct}}^\top \hat{\phi}_{\mathrm{struct}}
    +
    \bigl(1-\alpha(x)\bigr)
    \left(
        \mathbf{w}_{\mathrm{coh}}^\top \hat{\phi}_{\mathrm{coh}}
        +
        \mathbf{w}_{\mathrm{cont}}^\top \hat{\phi}_{\mathrm{cont}}
    \right),
\]
where
\[
    \alpha(x)
    =
    \exp\!\left(
        -\frac{1}{2\sigma^2}
        \bigl(\bar{w}-\mu\bigr)^2
    \right).
\]
Here $\hat{\phi}_{b}$ is the robust-scaled feature vector for block $b$, and $\bar{w}$ is the raw \textsc{WordsPerStep} value before scaling.
When $\bar{w}$ lies near the prose-heavy regime centred at $\mu$, $\alpha(x)$ approaches one, suppressing the coherence and content terms.
When $\bar{w}$ is far from this regime, $\alpha(x)$ approaches zero and all three blocks contribute.
Implementation details for step extraction, hedge counting, entity extraction, and slope computation are provided in Appendix~\ref{app:features}.

Table~\ref{tab:requirements} summarises the external components required by each detector.
This comparison is not the source of TRACT's accuracy; it clarifies its practical setting.
Once sampled traces are available, TRACT is a text-only trajectory scorer and does not require an auxiliary NLI model, embedding model, answer parser, or access to model internals.
\begin{table*}[!ht]
\centering
\small
\setlength{\tabcolsep}{6pt}
\begin{tabular}{lcccccccc}
\toprule
 & EMR & NCP & BSC & NCS & NSN & RACE & SEU & TRACT \\
\midrule
\textbf{NLI Model}       
&  & \checkmark &  &  & \checkmark  & \checkmark &  &  \\
\textbf{Embedding Model}  
&  &  & \checkmark & \checkmark & \checkmark & \checkmark & \checkmark &  \\
\bottomrule
\end{tabular}
\caption{\textbf{Method requirements.}
A checkmark indicates that a detector requires the corresponding external component.
TRACT operates directly on sampled reasoning trajectories and does not require external NLI or embedding models.}
\label{tab:requirements}
\end{table*}

Unlike NLI- and embedding-based baselines, TRACT computes all features directly from surface trajectory statistics.
This makes the scorer inexpensive to run, transparent to inspect, and compatible with closed-source models where only generated text is available. Figure~\ref{fig:block_ablation} evaluates whether TRACT's three feature blocks contribute complementary signal.
The full scorer is strongest or near-strongest across benchmarks, indicating that coherence, structure, and content capture different failure signatures rather than redundant variants of the same cue.

\section{Experiments}
\paragraph{Benchmarks}
We evaluate on four diverse reasoning benchmarks spanning distinct reasoning modalities and difficulty regimes: BBH-Tracking~\cite{suzgun2022challenging}, which requires multi-step state tracking under sequential object permutations; GPQA Diamond~\cite{rein2024gpqa}, which tests graduate-level scientific reasoning resistant to surface-level retrieval; MATH-500~\cite{lightman2023lets}, which covers competition-level mathematics across seven subject areas where intermediate step quality is predictive of correctness; and CausalT5K~\cite{causalt5k2026}, which probes structural causal reasoning under adversarial narrative pressure across Pearl's three rungs. For evaluation on CausalT5k, we leverage the D6 setting, encompassing environment and climate issues, under all three rungs. Together, these benchmarks exercise TRACT's trace-level features across symbolic, scientific, mathematical, and causal reasoning domains (for additional details see Appendix~\ref{app:datasets}).

\paragraph{Models}
For each benchmark, we sample responses from a mixture of open-weight and proprietary large language models spanning several model families: Nemotron-30B~\citep{nvidia2025nemotron30b}, GPT-OSS-120B~\citep{openai2025gptoss}, LLaMA-3-70B~\citep{touvron2024llama3}, Amazon Nova Pro~\citep{amazon2025nova}, and Gemma-3-27B~\citep{team2025gemma3}.
Additional model details, prompting, sampling hyperparameters, and step-extraction procedures are provided in Appendix~\ref{app:models}.

\paragraph{Scorers}
We compare against seven black-box UQ scorers with recommended settings: Exact-Match Repetition (EMR)~\citep{JMLR:v27:25-1557} , NLI-based Non-Contradiction Probability (NCP)~\citep{Manakuletal2023}, BERTScore Consistency (BSC)~\citep{zhang2020bertscore}, Normalized Cosine Similarity (NCS)~\citep{reimers2019sentencebert}, Normalized Semantic Negentropy (NSN)~\citep{farquhar2024semantic,JMLR:v27:25-1557}, RACE~\citep{Wang_Su_Ai_Liu_2026}, and Semantic Embedding Uncertainty (SEU)~\citep{grewal2024improvinguncertaintyquantificationlarge}.
All methods receive the same $10$ sampled traces.
We evaluate TRACT with $\mu=28$ words per step and $\sigma^2=50$.
Following prior work~\citep{Wang_Su_Ai_Liu_2026,kuhn2023semantic}, we report AUC as the primary metric.
Details of TRACT features are given in Table~\ref{tab:tract_features} and Appendix~\ref{app:features}.

In contrast to prior methods that 
rely heavily on final answer signatures to assess correctness, we design 
controlled interventions to isolate trajectory-level signals from answer-level artifacts. Specifically, 
we restrict evaluation to sets of sampled traces where the final answer is no longer a valid separator between incorrect and correct responses, ensuring that variation in scores arises from differences in reasoning structure, rhetorical coherence, and cross-trace 
consistency rather than answer disagreement. 

We introduce two complementary conditions. In 
\textsc{Force}, we replace each response's final answer with the ground truth and 
canonicalize the announcement of the final answer, producing trajectories that have final answers that are correct and uniform; this isolates whether TRACT detects inconsistencies independent of answer correctness. 
Additionally, in \textsc{Remove}, we strip explicit answer-announcement steps, preserving natural reasoning trajectories while eliminating explicit answer signals. Responses maintain their original label and performance is based on the initial response.
Together, these settings test whether incorrectness is reflected in the internal dynamics of reasoning 
trajectories, rather than the terminal output, providing a stricter evaluation of trajectory-level uncertainty signals.
\subsection{Results}

\begin{table*}[h]
\centering
\footnotesize
\setlength{\tabcolsep}{4pt}
\resizebox{\textwidth}{!}{%
\begin{tabular}{llccccc}
\toprule
Dataset & Method & Gemma-3-27B & GPT-OSS-120B &  LLaMA-3-70B & Nemotron-30B & Nova-Pro \\
\midrule
\multirow{8}{*}{BBH-Tracking} & EMR & 50.00 / 59.72 & 50.00 / 50.00 & 50.00 / 50.50 & 50.00 / 60.80 & 50.00 / 34.72 \\
 & NCP & 67.02 / 68.01 & 51.00 / 52.22 & 45.66 / 47.23 & 55.90 / 52.73 & 45.35 / 44.93 \\
 & BSC & 53.41 / 54.42 & 65.48 / \textbf{66.84} & 49.91 / 47.69 & 75.08 / 76.51 & 49.77 / \textbf{48.93} \\
 & NCS & 57.04 / 59.03 & 58.91 / 61.01 & 50.01 / 47.29 & 51.45 / 48.36 & 41.99 / 42.69 \\
 & NSN & 50.00 / 62.19 & 50.00 / 50.21 & 50.00 / 47.85 & 50.00 / 60.80 & 50.00 / 34.57 \\
 & RACE & 76.17 / 54.67 & \textbf{68.20} / 60.42 & 48.30 / 52.48 & 27.56 / 76.16 & 40.91 / 46.64 \\
 & SEU & 50.81 / 58.12 & 51.96 / 61.93 & 47.85 / 48.87 & 31.62 / 61.68 & \textbf{62.02} / 39.26 \\
\cmidrule(lr){2-7}
 & \textbf{TRACT} & \textbf{77.03} / \textbf{77.03} & 53.63 / 53.63 & \textbf{60.38} / \textbf{60.38} & \textbf{84.05} / \textbf{84.05} & 44.59 / 44.59 \\
\midrule
\multirow{8}{*}{GPQA Diamond} & EMR & 50.00 / 50.00 & 50.00 / 50.00 & 50.00 / 51.14 & 50.00 / 50.00 & 50.00 / 50.00 \\
 & NCP & 56.29 / 55.59 & 46.58 / 47.33 & 50.87 / 50.19 & 32.47 / 33.51 & 51.38 / 51.41 \\
 & BSC & 64.48 / 64.46 & 55.62 / 55.57 & 48.67 / 50.25 & 45.60 / 46.22 & 53.86 / 53.59 \\
 & NCS & 60.92 / 60.79 & 54.96 / 56.36 & 48.34 / 48.29 & 47.47 / 46.85 & 56.47 / 56.25 \\
 & NSN & 50.00 / 50.00 & 50.00 / 49.48 & 50.00 / 50.46 & 50.00 / 50.00 & 50.00 / 50.00 \\
 & RACE & 24.28 / 59.01 & 33.58 / 48.74 & 48.38 / 45.85 & 60.92 / 50.68 & 48.00 / 56.30 \\
 & SEU & 49.82 / 59.61 & 54.48 / 56.89 & 49.37 / 47.25 & 52.57 / 49.20 & 47.55 / 55.75 \\
\cmidrule(lr){2-7}
 & \textbf{TRACT}& \textbf{73.93} / \textbf{73.93} & \textbf{71.89} / \textbf{72.91} & \textbf{63.66} / \textbf{63.70} & \textbf{74.26} / \textbf{78.98} & \textbf{56.75} / \textbf{56.75} \\
\midrule
\multirow{8}{*}{Math 500} & EMR & 50.00 / 55.44 & 50.00 / 58.04 & 50.00 / 53.80 & 50.00 / 57.98 & 50.00 / 50.99 \\
 & NCP & 70.18 / 70.11 & 65.14 / 65.46 & 63.98 / 64.60 & 71.43 / 73.09 & 68.25 / 67.39 \\
 & BSC & 59.06 / 58.40 & 63.68 / 68.69 & 58.72 / 63.39 & 66.53 / 69.04 & 64.38 / 62.72 \\
 & NCS & 69.14 / 69.46 & 67.82 / 73.14 & 63.58 / 70.17 & \textbf{75.96} / 76.87 & 71.49 / 71.44 \\
 & NSN & 50.00 / 61.32 & 50.00 / \textbf{77.40} & 50.00 / 60.73 & 50.00 / \textbf{79.72} & 50.00 / 54.00 \\
 & RACE & 36.88 / 70.02 & 56.87 / 63.81 & 63.98 / \textbf{75.94} & 59.21 / 68.94 & 73.42 / 72.19 \\
 & SEU & 59.28 / 68.81 & 64.79 / 70.24 & 53.98 / 70.41 & 53.14 / 70.51 & 63.37 / 69.13 \\
\cmidrule(lr){2-7}
 & \textbf{TRACT} & \textbf{79.68} / \textbf{79.63} & \textbf{69.35} / 69.73 & \textbf{67.46} / 75.60 & 67.06 / 62.69 & \textbf{74.91} / \textbf{81.30} \\
\midrule
\multirow{8}{*}{CasualT5k-D4} & EMR & 50.00 / 50.00 & 50.00 / 50.00 & 50.00 / 50.00 & 50.00 / 38.59 & 50.00 / 50.00 \\
 & NCP & 50.00 / 44.71 & 50.00 / 59.46 & 50.00 / 42.30 & 50.00 / 66.79 & 50.00 / 39.70 \\
 & BSC & 44.96 / 45.74 & 54.03 / 61.91 & 54.87 / 49.07 & 65.45 / 36.45 & 50.54 / 45.71 \\
 & NCS & 48.62 / 50.37 & 44.22 / 60.09 & 54.52 / 46.01 & 58.48 / 46.79 & 53.36 / 47.40 \\
 & NSN & 50.00 / 51.42 & 50.00 / 39.27 & 50.00 / 50.52 & 50.00 / 38.88 & 50.00 / 53.64 \\
 & RACE & 60.60 / \textbf{64.22} & 60.00 / 55.72 & 56.44 / 52.90 & 67.40 / 66.90 & 50.58 / 53.22 \\
 & SEU & 45.95 / 55.46 & 48.87 / 63.12 & 57.35 / 45.25 & 62.38 / 49.68 & 44.30 / 47.55 \\
\cmidrule(lr){2-7}
 & \textbf{TRACT} & \textbf{63.01} / 63.01 & \textbf{70.34} / \textbf{70.34} & \textbf{59.96} / \textbf{59.96} & \textbf{75.44} / \textbf{75.44} & \textbf{56.45} / \textbf{56.45} \\
\bottomrule
\end{tabular}%
}
\caption{AUC (\%) across datasets. Each cell is reported as \textsc{Force} / \textsc{Remove}.
For each dataset and model, the best \textsc{Force} value and the best \textsc{Remove} value 
across methods are bolded.}
\label{tab:auc_v5_combined}
\end{table*}

\paragraph{Current detectors are susceptible to sanity checks}
A reliable uncertainty detector should evaluate how the model reasons, not merely what answer it gives.
To test this, we apply two oracle patching modes: \textsc{Force}, which replaces the final answer with the ground truth, and \textsc{Remove}, which deletes explicit answer-announcement steps.
A robust trace-level detector should remain informative under both interventions; large shifts between the two conditions indicate sensitivity to answer presentation rather than reasoning trajectory.

Table~\ref{tab:auc_v5_combined} shows that several existing detectors are unstable under these sanity checks.
For example, RACE achieves 76.17 AUC under \textsc{Force} on BBH-Tracking with Gemma-3-27B, but drops to 54.67 under \textsc{Remove}.
On GPQA Diamond with the same model, it nearly inverts, scoring 24.28 under \textsc{Force} and 59.01 under \textsc{Remove}.
NSN exhibits a different failure mode on MATH-500: it remains at chance under \textsc{Force} for every model, yet rises sharply under \textsc{Remove}, reaching 77.40 for GPT-OSS-120B and 79.72 for Nemotron-30B.
These shifts suggest that the detectors are responding strongly to endpoint formatting or answer availability rather than to stable reasoning evidence.

TRACT is more stable across the same interventions.
In Table~\ref{tab:auc_v5_combined}, TRACT often obtains identical or near-identical \textsc{Force}/\textsc{Remove} scores because its features are computed from the reasoning body after answer-announcement removal.
It also remains competitive in absolute AUC, especially on settings where surface answer cues are weak.
On CausalT5K-D6, for instance, TRACT obtains 75.44/75.44 with Nemotron-30B and 70.34/70.34 with GPT-OSS-120B, while many baselines remain close to chance or vary substantially across intervention modes.
Together with Figure~\ref{fig:stability_plot}, these results indicate that TRACT is less dependent on answer-level artifacts and more sensitive to trajectory-level signals.

\begin{figure}[t]
  \centering
  \begin{minipage}[t]{0.48\linewidth}
    \centering
    \includegraphics[width=\linewidth]{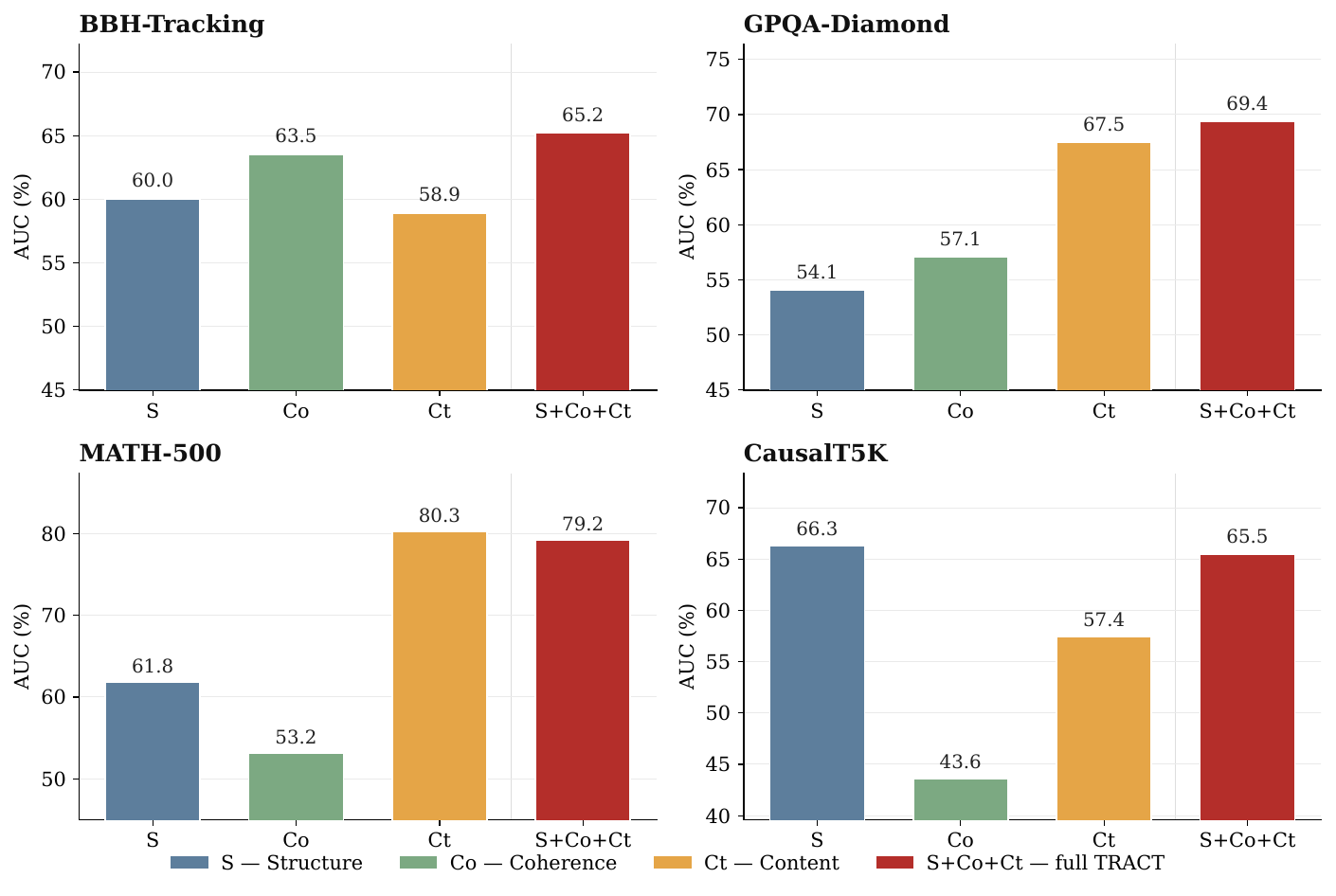}
    \captionof{figure}{Ablating TRACT feature blocks across four benchmarks. \textbf{S} = Structure, \textbf{Co} = Coherence, and \textbf{Ct} = Content. The full S+Co+Ct scorer is strongest or near-strongest across benchmarks, indicating that the blocks capture complementary trace-level signals.}
    \label{fig:block_ablation}
  \end{minipage}\hfill
    \begin{minipage}[t]{0.50\linewidth}
    \centering
    \includegraphics[width=\linewidth]{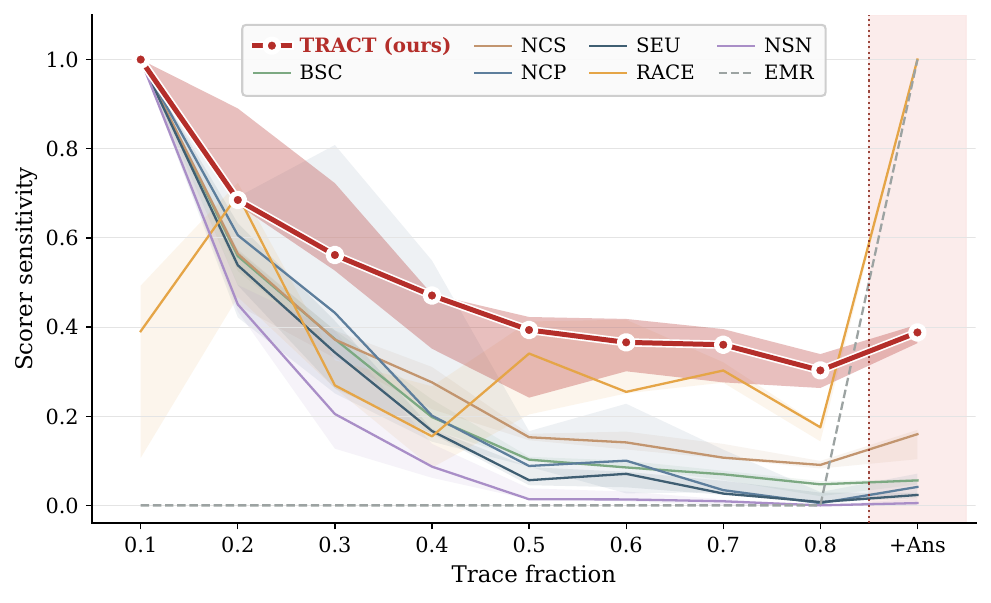}
    \captionof{figure}{\textbf{Step-wise sensitivity on GPQA Diamond.}
Normalised score changes as progressively more of the trace is revealed. Endpoint-reliant methods spike when the final answer appears; TRACT remains sensitive across the reasoning body.}
    \label{fig:sensitivity}
  \end{minipage}
\end{figure}

\paragraph{Step-wise sensitivity analysis}
Figure~\ref{fig:sensitivity} examines where each detector obtains its signal along the reasoning trajectory.
For each response, we reveal progressively larger prefixes of the trace and measure the marginal change in detector score after each additional step.
Scores are normalised to $[0,1]$ within each method so that the figure compares where sensitivity occurs rather than raw score magnitude.

The methods separate into three qualitative patterns.
First, answer-anchored methods such as EMR and RACE remain relatively insensitive throughout the reasoning body and respond most strongly when the final answer is revealed.
This behaviour is undesirable for a reasoning-aware detector: it indicates that the method is primarily reading the endpoint. 

Second, methods such as NCS, NCP, BSC, SEU, and NSN are prone to saturating, and react strongly to early steps and then plateau.
These methods extract useful information from the trace, but much of their signal is concentrated near the beginning rather than distributed across the full trajectory.

TRACT behaves differently.
Its sensitivity remains more evenly distributed across the reasoning process and does not spike primarily at the answer-reveal transition.
This reflects the design of Table~\ref{tab:tract_features}: TRACT combines static coherence, dynamic structural trends, and cross-sample content agreement, so no single step or endpoint dominates the score.

\paragraph{Dataset and feature relationships}
Figure~\ref{fig:block_ablation} decomposes TRACT by feature block and shows that different benchmarks expose different trace-level failure modes.
On MATH-500 and GPQA Diamond, content agreement is especially informative: the Content block alone reaches 80.3 AUC on MATH-500 and 67.5 on GPQA Diamond.
This suggests that, for mathematical and scientific reasoning tasks, incorrect sampled traces often diverge in the concepts or final vocabulary they converge to.

CausalT5K-D6 shows the opposite pattern.
There, the Structure block is strongest among individual blocks, reaching 66.3 AUC.
This is consistent with the nature of causal reasoning under adversarial narrative pressure: correctness depends less on repeating the same lexical endpoint and more on maintaining a coherent reasoning trajectory.

The full block-combination heatmap in Appendix~\ref{app:features} (Figure~\ref{fig:block_heatmap}) further shows that the blocks are complementary rather than redundant.
Structure plus Content is the strongest two-block combination on GPQA Diamond and CausalT5K-D6, while on MATH-500 it nearly matches Content alone.
The full TRACT scorer, combining Structure, Coherence, and Content, achieves the highest or near-highest AUC on every benchmark.
This supports the design choice in Table~\ref{tab:tract_features}: TRACT does not rely on a single universal cue, but combines several weak, interpretable signatures whose usefulness varies by task.

\paragraph{Complementarity with existing detectors}
Finally, we test whether TRACT contributes signal beyond existing black-box scorers.
Appendix~\ref{app:pairwise} fuses TRACT with each baseline using 4-fold cross-validated logistic regression on unmodified traces.
This is a diagnostic rather than our proposed deployment method: improvements indicate that TRACT contains signal not already captured by the partner scorer.

\begin{table}[!ht]
\centering
\small
\setlength{\tabcolsep}{5pt}
\caption{\textbf{TRACT is complementary to existing detectors.}
Average AUC across four benchmarks for standalone TRACT and the best TRACT+baseline fusion.}
\label{tab:pairwise_main}
\begin{tabular}{@{}lccc@{}}
\toprule
\textbf{Model} & \textbf{TRACT} & \textbf{Best TRACT+X} & \textbf{Gain} \\
\midrule
Nova Pro       & 59.05 & 79.05 & +20.00 \\
Gemma-3-27B    & 74.59 & 80.01 & +5.42 \\
GPT-OSS-120B   & 71.53 & 84.10 & +12.57 \\
LLaMA-3-70B    & 64.65 & 74.72 & +10.07 \\
Nemotron-30B   & 78.38 & 88.31 & +9.93 \\
\bottomrule
\end{tabular}
\end{table}

Fusion improves over standalone TRACT for every model, with average gains from +5.42 to +20.00 AUC points.
EMR and NSN are the most consistent partners, showing that TRACT complements both answer-repetition and semantic-entropy signals.
RACE also helps for GPT-OSS-120B and Nemotron-30B, suggesting that TRACT's hedge, step-length, and lexical-convergence features are not fully captured by reasoning-consistency or answer-alignment scores.
\section{Conclusion}
\label{sec:conclusion}
 
We have presented a controlled-invariance framework for evaluating whether hallucination detection methods genuinely assess reasoning trajectories or rely on answer-level artifacts.
Our \textsc{Force} and \textsc{Remove} oracle tests provide a simple, model-agnostic diagnostic: when the reasoning body is preserved, a reasoning-faithful detector should remain informative even if the final answer is forced to the ground truth or explicit answer-announcement steps are removed.
Applying these tests across five models and four benchmarks, we find that many current methods are more sensitive to endpoint cues than standard evaluations reveal.
 As long-form reasoning becomes a defining capability of frontier models, and as hallucination detection matures from proof-of-concept into a relied-upon safety mechanism, the standards by which we evaluate these methods must keep pace.
Reporting performance on unperturbed traces is insufficient if that performance collapses, or persists for the wrong reasons, under controlled perturbation.
We encourage future work in this area to adopt oracle robustness testing as a standard evaluation practice, ensuring that progress in hallucination detection reflects genuine advances in reasoning assessment rather than increasingly sophisticated exploitation of surface cues.

\paragraph{Limitations}
Force and Remove should be understood as necessary sanity checks, not sufficient guarantees of reasoning fidelity. 
Passing these tests does not prove that a detector verifies intermediate logical steps; it only rules out certain forms of answer-level dependence. 
TRACT inherits this limitation by design: its lexical and structural features are interpretable and efficient, but they are not a semantic proof of correctness. 
Our evaluation is limited to benchmark tasks with canonical answers;  
generalizing the framework to open-ended long-form generation, tool use, and domains where correctness is not reducible to final answers requires additional intervention designs.
\newpage
\bibliographystyle{plainnat}
\bibliography{references}







\appendix

\section{Datasets}
\label{app:datasets}
 
We describe the four benchmarks used to evaluate TRACT and the other black-box scorers.
 
\subsection{BBH-Tracking (BIG-Bench Hard)}
 
\textbf{BIG-Bench Hard} (BBH; \citealt{suzgun2022challenging}) is a curated subset of 23 tasks drawn from the broader BIG-Bench evaluation suite, selected because prior language model evaluations failed to exceed average human-rater performance on them under standard few-shot prompting. BBH spans a diverse range of reasoning skills---multi-step algorithmic reasoning, logical deduction, natural language understanding, and commonsense inference---with 250 examples per task (6{,}511 total). Each task includes both answer-only and chain-of-thought (CoT) few-shot prompt sets. A central finding of the original paper is that CoT prompting substantially closes the human-model gap, enabling Codex to surpass human-rater performance on 17 of the 23 tasks.
 
We focus on the \textbf{Tracking Shuffled Objects} subtask. Each problem specifies $N$ agents each initially holding one of $N$ distinct objects, then narrates $N$ pairwise swaps; the model must identify which object a designated agent holds at the end. Problems are constructed so that every agent participates in at least one swap and no two agents swap back-to-back, ruling out trivial surface shortcuts. Correct solutions require maintaining an explicit, incrementally updated state across multiple reasoning steps, which makes the coherence and rhetorical structure of the chain of thought directly diagnostic of solution quality.
 
\subsection{GPQA Diamond}
 
\textbf{GPQA} (Graduate-Level Google-Proof Q\&A; \citealt{rein2024gpqa}) is a benchmark of 448 four-way multiple-choice questions written by domain experts in biology, physics, and chemistry. Questions are deliberately designed to be ``Google-proof'': highly skilled non-expert validators reach only 34\% accuracy despite an average of over 30 minutes of unrestricted web access. Domain experts holding or pursuing PhDs in the relevant fields achieve 65\% accuracy (74\% after discounting questions the experts themselves identified as flawed in retrospect).
 
\textbf{GPQA Diamond} is the hardest curated subset of 198 questions, retaining only those for which both independent expert annotators answered correctly while the majority of non-expert validators answered incorrectly. The random-chance baseline on Diamond is 25\%. This subset is now a standard frontier evaluation: at the time of the benchmark's release, GPT-4 achieved 39\%; subsequent models have improved substantially, but the dataset remains a meaningful discriminator for state-of-the-art systems. For TRACT evaluation, GPQA Diamond provides a pool of problems demanding deep, multi-step scientific reasoning where superficial lexical cues are insufficient and the internal structure of a reasoning trace is strongly predictive of correctness.
 
\subsection{MATH-500}
 
\textbf{MATH-500} \citep{lightman2023lets} is a 500-problem subset of the MATH dataset \citep{hendrycks2021measuring}, curated by OpenAI to support evaluation of process supervision methods for mathematical reasoning. The full MATH dataset comprises 12{,}500 competition-style problems drawn from AMC 10, AMC 12, AIME, and related contests; MATH-500 samples representatively from seven subject areas---Prealgebra, Algebra, Number Theory, Counting and Probability, Geometry, Intermediate Algebra, and Precalculus---across five difficulty levels (1--5). Each problem includes a complete step-by-step reference solution in \LaTeX.
 
MATH-500 was introduced to compare outcome supervision (feedback on final answers only) against process supervision (feedback on each intermediate reasoning step), with process supervision found to significantly outperform outcome supervision. This is precisely the regime where the quality of intermediate reasoning steps, rather than the final answer alone, determines success. MATH-500 therefore provides a natural evaluation bed for TRACT's trace-level coherence features: problems are hard enough that flawed intermediate steps frequently co-occur with incorrect final answers, making correct step-by-step structure a reliable signal of correctness.
 
\subsection{CausalT5K}
 
\textbf{CausalT5K} \citep{causalt5k2026} is a diagnostic benchmark of over 5{,}000 cases (5{,}147 total) spanning 10 domains, designed to stress-test three distinct failure modes in LLM causal reasoning: (1)~\emph{rung collapse}, where a model responds to an interventional or counterfactual query with associational evidence, conflating different levels of Pearl's causal hierarchy; (2)~\emph{sycophantic drift}, where a model abandons a correct causal claim under adversarial rhetorical pressure; and (3)~\emph{miscalibrated refusal}, where a model either refuses valid causal claims (the Skepticism Trap) or endorses underdetermined ones.
 
Unlike purely synthetic causal benchmarks, CausalT5K embeds causal traps in realistic narrative scenarios across domains including medicine, economics, history, sports, and daily life, developed through a rigorous human-machine collaborative pipeline involving 40 domain experts. Performance is decomposed into a two-axis scheme of Utility (sensitivity---correctly affirming valid causal claims) and Safety (specificity---correctly rejecting invalid ones), revealing failure modes invisible to aggregate accuracy. CausalT5K spans all three of Pearl's rungs (associational $\mathcal{L}_1$, interventional $\mathcal{L}_2$, counterfactual $\mathcal{L}_3$) and is unique among existing causal benchmarks in annotating trap type, applying adversarial pressure, and reporting the two-axis decomposition. For TRACT, this dataset provides a qualitatively distinct reasoning challenge relative to the other benchmarks: correct responses require not only multi-step deduction but also resistance to misleading narrative framing, making trace-level rhetorical and coherence features especially salient.

\section{Model Configurations and Sampling Procedures}
\label{app:models}
 
\subsection{Models}
 
We evaluate across five models spanning four LLM families and a range of parameter scales, mixing open-weight and proprietary systems to reduce the risk of findings that are idiosyncratic to a single architecture or training pipeline.
Table~\ref{tab:models} summarises the key properties of each model.
 
\begin{table}[ht]
\centering
\small
\caption{Models used in our evaluation. All models are instruction-tuned variants. Parameter counts are approximate and reflect the publicly reported size of each model.}
\label{tab:models}
\begin{tabular}{@{}llll@{}}
\hline
\textbf{Model} & \textbf{Family} & \textbf{Parameters} & \textbf{Access} \\
\hline
Nemotron-30B~\citep{nvidia2025nemotron30b}   & Nemotron (NVIDIA)   & 30B   & Open weights \\
GPT-OSS-120B~\citep{openai2025gptoss}       & GPT (OpenAI)        & 120B  & Open weights \\
LLaMA-3-70B~\citep{touvron2024llama3}        & LLaMA (Meta)        & 70B   & Open weights \\
Amazon Nova Pro~\citep{amazon2025nova}        & Nova (Amazon)       & ---   & API \\
Gemma-3-27B~\citep{team2025gemma3}           & Gemma (Google)      & 27B   & Open weights \\
\hline
\end{tabular}
\end{table}
 
\subsection{Prompting}
 
All models are prompted with a standardised step-by-step instruction that elicits chain-of-thought reasoning before a final answer.
The system prompt is:
 
\begin{quote}
\small
\texttt{Solve the following problem step by step. Show your reasoning at each step, then provide your final answer on the last line in the format: "Final Answer: <answer>".}
\end{quote}
 
We do not use few-shot exemplars; all evaluations are zero-shot.
The prompt is identical across all models for BBH-Tracking, Math-500, and GPQA-Diamond to ensure that variation in reasoning traces reflects model behaviour rather than prompt engineering.

For CausalT5k, we utilize the prompt:
\begin{quote}
\small
\texttt{You are evaluating a causal claim. Read the scenario and claim carefully, then reason through your answer step by step.\\[0.5em]
Scenario: \{scenario\}\\[0.3em]
Claim: \{claim\}\\[0.3em]
Variables:\\
- X (exposure): \{variables.X.name\}\\
- Y (outcome): \{variables.Y.name\}\\
- Z (confounders/other factors): \{variables.Z\}\\[0.5em]
Your task: Evaluate whether the claim is causally valid. Think through your reasoning step by step, then provide your answer.\\[0.5em]
Step 1: Identify the causal relationship being claimed.\\[0.3em]
Step 2: Consider potential confounders or alternative explanations.\\[0.3em]
Step 3: Evaluate whether the evidence supports a causal relationship.\\[0.3em]
Step 4: Determine if additional information is needed to make a reliable judgment.\\[0.3em]
Step 5: State your conclusion.\\[0.5em]
Final Answer: [YES, NO, or REFUSE]\\[0.3em]
Explanation:}
\end{quote}

\subsection{Sampling}
 
For each (model, benchmark) pair, we sample $K$ independent reasoning traces per prompt using nucleus sampling.
Table~\ref{tab:sampling} reports the sampling hyperparameters.
 
\begin{table}[ht]
\centering
\caption{Sampling hyperparameters. All models use the same configuration except where noted.}
\label{tab:sampling}
\begin{tabular}{@{}ll@{}}
\hline
\textbf{Parameter} & \textbf{Value} \\
\hline
Responses per prompt ($K$)  & 10 \\
Temperature                 & 0.3 \\
Top-$p$ (nucleus)           & 0.95 \\
Max output tokens           & 2048 \\
\hline
\end{tabular}
\end{table}
 
\subsection{Step Extraction}
 
Reasoning traces are segmented into steps by splitting on double newlines.
When a model produces unstructured output (e.g.\ a single block of prose with no paragraph breaks), we fall back to splitting on numbered or bulleted list boundaries and, as a last resort, single newlines.
Steps shorter than 5 characters or consisting entirely of punctuation and markdown formatting are discarded.
Steps matching answer-announcement patterns (e.g.\ ``Final Answer:'', ``The answer is'') are removed from the trace body before feature extraction to prevent answer-level artifacts from leaking into trajectory features.
 
\subsection{Infrastructure}
 
Open-weight models are served via vLLM~\citep{kwon2023vllm} on NVIDIA A100 GPUs with tensor parallelism as needed (TP$=$2 for LLaMA-3-70B, TP$=$1 for all others).
Amazon Nova Pro is accessed through the Amazon Bedrock API.
All inference is run with \texttt{TRANSFORMERS\_OFFLINE=1} using locally cached model weights.

\section{Feature Inventory}
\label{app:features}
\begin{figure}[t]
    \centering
    \includegraphics[width=.95\linewidth]{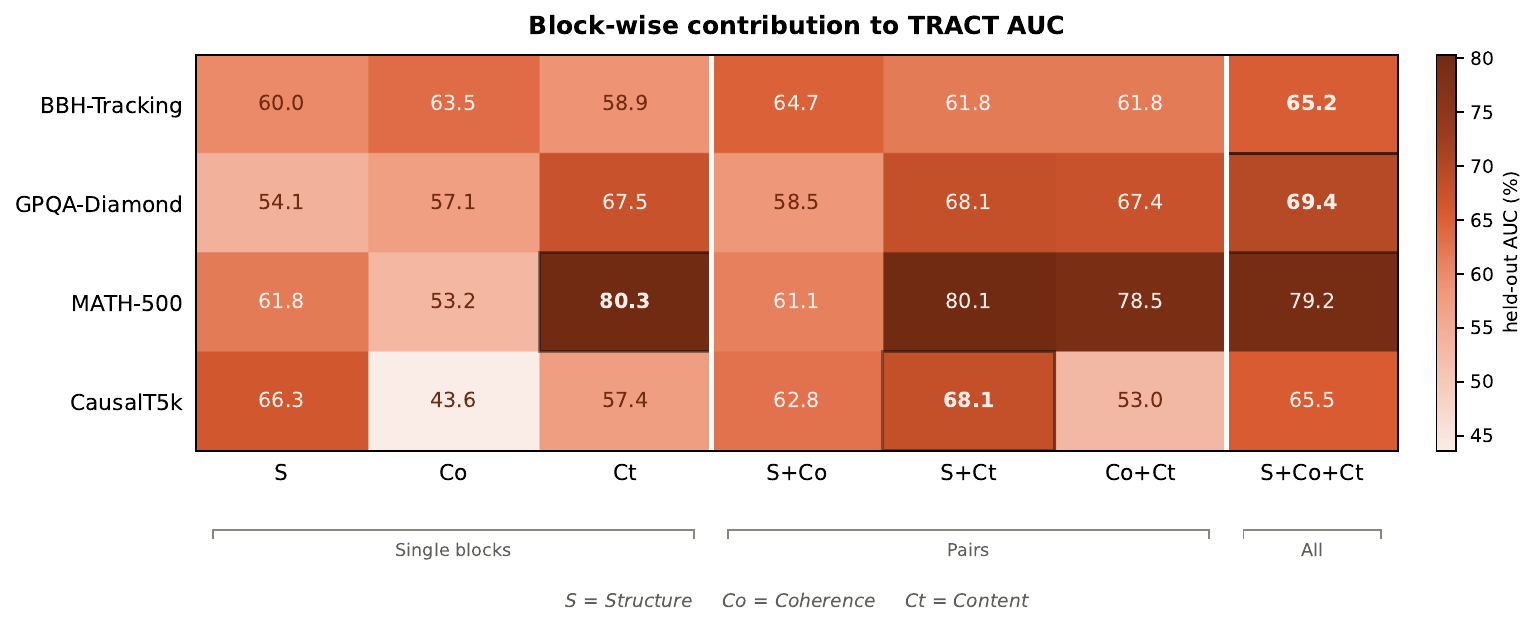}
    \caption{Block-wise contribution to TRACT AUC across four benchmarks, averaged over five models. S = Structure (cross-response rhetorical and structural signals), Co = Coherence, Ct = Content (cross-response lexical agreement). White dividers separate single-block ablations, pairwise combinations, and the full scorer. Bold outlines mark the highest-scoring subset per dataset.}
    \label{fig:block_heatmap}
\end{figure}
Table~\ref{tab:tract_features} in the main paper gives the complete TRACT feature inventory, including the diagnostic intuition, sign, and implementation-level definition for each feature.
Here we provide additional implementation details.

For features defined independently within a trace, we compute the statistic for each of the $K$ sampled traces and then average across traces.
For cross-trace divergence features, we compute the average pairwise Jaccard distance across all unordered trace pairs.

The hedge count $h^{(k)}_i$ is computed using a fixed lexicon $\mathcal{H}$ containing uncertainty and contrast markers such as \emph{however}, \emph{although}, \emph{maybe}, \emph{perhaps}, \emph{might}, \emph{could}, \emph{seems}, and \emph{hmm}.
The entity set $\mathcal{E}^{(k)}_i$ is approximated by capitalised tokens, excluding sentence-initial function words and answer-formatting tokens.
Slopes are ordinary least-squares slopes against the normalised step position $i/T_k$ within each trace.

Before scoring, each feature is robust-scaled using median centring and IQR normalisation, then clipped to $[-3,3]$.
The structure block always contributes to the final score.
The coherence and content blocks are modulated by the verbosity gate described in Section~\ref{sec:tract}.


\section{TRACT Pairwise Complementarity}
\label{app:pairwise}

The main paper evaluates TRACT as a standalone detector.
Here we ask a diagnostic question: does TRACT capture information that is complementary to existing black-box uncertainty scorers?
To test this, we fuse TRACT with each baseline scorer using a 4-fold cross-validated logistic regression classifier.
All fusion experiments use the unmodified trace setting, class-weighted logistic regression, standardised features, and regularisation parameter $C=1.0$.

This analysis is not intended to propose an ensemble as the primary method.
Instead, it tests whether TRACT contains signal that is not already captured by each baseline.
Improvement over EMR indicates information beyond answer repetition; improvement over NCS, BSC, or SEU indicates information beyond embedding or semantic similarity; improvement over NCP or NSN indicates information beyond entailment or semantic-entropy-style uncertainty; and improvement over RACE indicates signal beyond reasoning-consistency and answer-alignment components.

Table~\ref{tab:pairwise_summary} summarises the main result.
Across all five models, the best TRACT+partner fusion improves over standalone TRACT, with average gains ranging from +5.42 to +20.00 AUC points across benchmarks.
The most consistently helpful partners are EMR and NSN, indicating that TRACT complements both simple answer-repetition signals and semantic-entropy-style uncertainty.
For GPT-OSS-120B and Nemotron-30B, RACE is also among the most useful partners, suggesting that TRACT's explicit step-level trajectory features provide signal not fully captured by aggregate reasoning-consistency or answer-alignment scores.

\begin{table}[H]
\centering
\small
\caption{\textbf{Summary of TRACT pairwise complementarity.}
For each model, we report standalone TRACT AUC averaged over the four benchmarks, the best average TRACT+partner fusion, and the average gain over standalone TRACT.
The final column lists the partners with the highest number of wins over standalone TRACT across benchmarks.}
\label{tab:pairwise_summary}
\begin{tabular}{@{}lcccc@{}}
\toprule
\textbf{Model} & \textbf{TRACT avg.} & \textbf{Best avg. fusion} & \textbf{Avg. gain} & \textbf{Most consistent partners} \\
\midrule
Nova-Pro       & 59.05 & 79.05 & +20.00 & EMR, NSN \\
Gemma-3-27B    & 74.59 & 80.01 & +5.42  & EMR, NSN, NCP \\
GPT-OSS-120B   & 71.53 & 84.10 & +12.57 & EMR, NSN, RACE \\
LLaMA-3-70B    & 64.65 & 74.72 & +10.07 & EMR, NCP, BSC, NSN \\
Nemotron-30B   & 78.38 & 88.31 & +9.93  & EMR, NSN, RACE \\
\bottomrule
\end{tabular}
\end{table}

The detailed results below show that complementarity is not confined to a single model or benchmark.
Nova-Pro benefits strongly from fusion, suggesting that TRACT supplies missing trajectory information when standalone performance is modest.
Gemma-3-27B and Nemotron-30B already have strong standalone TRACT scores, yet fusion still improves performance, indicating that TRACT is compatible with other uncertainty signals rather than simply replacing them.
For GPT-OSS-120B and LLaMA-3-70B, the best partner varies by dataset, which further supports the view that no single baseline subsumes TRACT's trajectory features.

In Tables~\ref{tab:fusion_nova}--\ref{tab:fusion_nemotron}, \textbf{bold} entries indicate fusion scores that improve over both standalone TRACT and the partner scorer on the corresponding dataset.
The final row, \textit{Wins vs. T}, counts the number of benchmarks, out of four, where TRACT+partner improves over standalone TRACT, regardless of whether it also beats the partner alone.

\subsection{Nova-Pro}

\begin{table}[H]
\centering
\scriptsize
\caption{\textbf{TRACT fusion results for Nova-Pro.}
Nova-Pro shows the largest complementarity gains overall.
EMR and NSN are the most consistent partners, indicating that TRACT contributes trajectory information that combines well with both endpoint repetition and semantic uncertainty.}
\label{tab:fusion_nova}
\resizebox{\textwidth}{!}{%
\begin{tabular}{@{}lrrrrrrrr@{}}
\toprule
\textbf{Dataset} & \textbf{T} & \textbf{+EMR} & \textbf{+NCP} & \textbf{+BSC} & \textbf{+NCS} & \textbf{+NSN} & \textbf{+SEU} & \textbf{+RACE} \\
\midrule
MATH-500      & 78.37 & \textbf{92.77} & \textbf{79.64} & 78.19 & \textbf{79.96} & \textbf{91.85} & \textbf{88.42} & \textbf{84.77} \\
GPQA Diamond  & 56.79 & \textbf{70.51} & 56.68 & 56.00 & 55.01 & \textbf{70.10} & \textbf{68.73} & 57.39 \\
BBH-Tracking  & 44.59 & 71.65 & 27.43 & 26.34 & 20.08 & 61.30 & 68.41 & 64.78 \\
CausalT5K-D4  & 56.45 & \textbf{81.27} & 54.70 & 52.66 & 50.30 & \textbf{79.66} & 53.41 & 51.91 \\
\midrule
\textit{Wins vs. T} & -- & 4/4 & 1/4 & 0/4 & 1/4 & 4/4 & 2/4 & 2/4 \\
\bottomrule
\end{tabular}%
}
\end{table}

\subsection{Gemma-3-27B}

\begin{table}[H]
\centering
\scriptsize
\caption{\textbf{TRACT fusion results for Gemma-3-27B.}
Standalone TRACT is already strong on Gemma-3-27B, but fusion still improves on most benchmarks.
The broad usefulness of EMR, NSN, and NCP suggests that trajectory features remain complementary even when answer-level and entailment-style signals are available.}
\label{tab:fusion_gemma}
\resizebox{\textwidth}{!}{%
\begin{tabular}{@{}lrrrrrrrr@{}}
\toprule
\textbf{Dataset} & \textbf{T} & \textbf{+EMR} & \textbf{+NCP} & \textbf{+BSC} & \textbf{+NCS} & \textbf{+NSN} & \textbf{+SEU} & \textbf{+RACE} \\
\midrule
MATH-500      & 82.97 & \textbf{86.30} & \textbf{83.40} & 82.87 & \textbf{83.10} & \textbf{87.03} & 93.84 & 93.56 \\
GPQA Diamond  & 75.36 & \textbf{77.92} & \textbf{75.55} & 74.66 & 75.23 & \textbf{79.16} & \textbf{77.73} & 75.29 \\
BBH-Tracking  & 77.03 & \textbf{77.21} & 76.84 & \textbf{78.51} & \textbf{78.42} & 75.53 & 74.69 & 74.55 \\
CausalT5K-D4  & 63.01 & \textbf{68.16} & \textbf{63.68} & 59.07 & 61.88 & \textbf{68.54} & 61.82 & \textbf{63.27} \\
\midrule
\textit{Wins vs. T} & -- & 4/4 & 3/4 & 1/4 & 2/4 & 3/4 & 2/4 & 2/4 \\
\bottomrule
\end{tabular}%
}
\end{table}

\subsection{GPT-OSS-120B}

\begin{table}[H]
\centering
\scriptsize
\caption{\textbf{TRACT fusion results for GPT-OSS-120B.}
Fusion improves strongly on BBH-Tracking and CausalT5K-D4, where trajectory instability and endpoint uncertainty provide complementary evidence.
The useful partners vary by dataset, indicating that TRACT is not subsumed by any single baseline family.}
\label{tab:fusion_gptoss}
\resizebox{\textwidth}{!}{%
\begin{tabular}{@{}lrrrrrrrr@{}}
\toprule
\textbf{Dataset} & \textbf{T} & \textbf{+EMR} & \textbf{+NCP} & \textbf{+BSC} & \textbf{+NCS} & \textbf{+NSN} & \textbf{+SEU} & \textbf{+RACE} \\
\midrule
MATH-500      & 81.32 & \textbf{83.09} & 81.21 & \textbf{81.62} & \textbf{82.44} & \textbf{81.45} & 89.93 & \textbf{90.23} \\
GPQA Diamond  & 80.82 & \textbf{81.00} & 79.17 & 80.32 & 80.36 & 80.71 & 86.99 & 85.55 \\
BBH-Tracking  & 53.63 & 72.90 & 46.27 & 57.65 & 51.06 & 76.53 & 57.64 & 52.18 \\
CausalT5K-D4  & 70.34 & \textbf{81.18} & \textbf{78.22} & \textbf{71.59} & \textbf{71.78} & \textbf{79.14} & \textbf{82.66} & \textbf{73.76} \\
\midrule
\textit{Wins vs. T} & -- & 4/4 & 1/4 & 2/4 & 2/4 & 2/4 & 3/4 & 3/4 \\
\bottomrule
\end{tabular}%
}
\end{table}

\subsection{LLaMA-3-70B}

\begin{table}[H]
\centering
\scriptsize
\caption{\textbf{TRACT fusion results for LLaMA-3-70B.}
For LLaMA-3-70B, EMR, NCP, BSC, and NSN frequently improve over standalone TRACT.
This pattern suggests that TRACT is most complementary to lexical, entailment, and semantic-entropy signals for this model, while SEU and RACE are less consistently helpful.}
\label{tab:fusion_llama}
\resizebox{\textwidth}{!}{%
\begin{tabular}{@{}lrrrrrrrr@{}}
\toprule
\textbf{Dataset} & \textbf{T} & \textbf{+EMR} & \textbf{+NCP} & \textbf{+BSC} & \textbf{+NCS} & \textbf{+NSN} & \textbf{+SEU} & \textbf{+RACE} \\
\midrule
MATH-500      & 75.35 & \textbf{91.59} & \textbf{77.11} & \textbf{75.38} & \textbf{77.66} & \textbf{89.32} & 85.46 & 78.96 \\
GPQA Diamond  & 62.90 & 62.83 & \textbf{63.23} & \textbf{63.81} & 62.42 & 62.63 & 60.97 & 62.09 \\
BBH-Tracking  & 60.38 & \textbf{79.46} & 51.23 & \textbf{60.70} & \textbf{63.18} & \textbf{78.32} & 76.11 & 63.30 \\
CausalT5K-D4  & 59.96 & \textbf{61.39} & \textbf{62.93} & 49.94 & 58.35 & \textbf{63.66} & 58.72 & 64.02 \\
\midrule
\textit{Wins vs. T} & -- & 3/4 & 2/4 & 3/4 & 2/4 & 2/4 & 1/4 & 3/4 \\
\bottomrule
\end{tabular}%
}
\end{table}

\subsection{Nemotron-30B}

\begin{table}[H]
\centering
\scriptsize
\caption{\textbf{TRACT fusion results for Nemotron-30B.}
TRACT remains complementary despite strong standalone performance.
EMR, NSN, and RACE improve on all four benchmarks, showing that TRACT's trajectory-level statistics combine well with both endpoint-based and reasoning-aware uncertainty signals.}
\label{tab:fusion_nemotron}
\resizebox{\textwidth}{!}{%
\begin{tabular}{@{}lrrrrrrrr@{}}
\toprule
\textbf{Dataset} & \textbf{T} & \textbf{+EMR} & \textbf{+NCP} & \textbf{+BSC} & \textbf{+NCS} & \textbf{+NSN} & \textbf{+SEU} & \textbf{+RACE} \\
\midrule
MATH-500      & 76.42 & \textbf{76.95} & \textbf{81.62} & \textbf{79.33} & \textbf{81.01} & \textbf{78.15} & 87.34 & \textbf{81.15} \\
GPQA Diamond  & 77.59 & \textbf{87.72} & 77.59 & 77.03 & \textbf{78.22} & \textbf{87.75} & 76.80 & \textbf{81.24} \\
BBH-Tracking  & 84.05 & \textbf{97.21} & 83.58 & \textbf{84.24} & 82.98 & \textbf{92.48} & \textbf{94.52} & \textbf{92.12} \\
CausalT5K-D4  & 75.44 & \textbf{80.95} & \textbf{76.05} & \textbf{78.45} & \textbf{76.07} & \textbf{80.65} & \textbf{75.45} & \textbf{79.66} \\
\midrule
\textit{Wins vs. T} & -- & 4/4 & 2/4 & 3/4 & 3/4 & 4/4 & 2/4 & 4/4 \\
\bottomrule
\end{tabular}%
}
\end{table}

Overall, the fusion analysis supports the interpretation of TRACT as a complementary trajectory scorer rather than a reparameterisation of existing uncertainty baselines.
The strongest partners differ across models and datasets, but the consistent gains over standalone TRACT indicate that its coherence, structure, and content features capture trace-level information that can be combined with answer repetition, semantic similarity, entailment, semantic entropy, and reasoning-consistency signals.


\newpage

\end{document}